\documentclass[11pt]{article}
\usepackage{coling2020}
\usepackage{times}
\usepackage{latexsym}
\usepackage{amsfonts}
\usepackage{xcolor}
\usepackage{footmisc}
\usepackage{dblfloatfix}
\usepackage{url}
\usepackage{wrapfig}

\usepackage{microtype}
\usepackage{amsmath}
\usepackage{amsthm}
\usepackage{textcomp}
\usepackage{mathtools}

\usepackage{comment}
\usepackage{multirow}
\usepackage{cleveref}
\crefformat{section}{\S#2#1#3}
\crefformat{subsection}{\S#2#1#3}
\crefformat{subsubsection}{\S#2#1#3}
\crefformat{appendix}{Appendix #2#1#3}
\crefformat{equation}{Eq.~(#2#1#3)}
\crefformat{table}{Table~#2#1#3}
\crefformat{figure}{Figure~#2#1#3}
\usepackage{booktabs}
\usepackage{enumitem}
\usepackage{graphicx}
\usepackage{subcaption}
\usepackage{wrapfig}

\usepackage{xcolor}
\definecolor{decentgrey}{RGB}{232,232,232}

\usepackage{tcolorbox}
\newtcbox{\greybox}{on line,colback=decentgrey,colframe=white,size=fbox,arc=3pt, box align=base, before upper=\strut, top=0pt, bottom=0pt, boxrule=0pt}

\definecolor{beaublue}{rgb}{0.4, 0.6, 0.8}
\definecolor{atomictangerine}{rgb}{1.0, 0.6, 0.4}

\usepackage{microtype}

\colingfinalcopy % Uncomment this line for the final submission

\title{DoLFIn: Distributions over Latent Features for Interpretability}

\author{Phong Le\thanks{Now at Amazon, Cambridge, UK.} \\
  Independent Researcher \\
  \texttt{lephong.xyz@gmail.com} \\ \And 
  Willem Zuidema \\
  University of Amsterdam \\
  \texttt{zuidema@uva.nl} \\}
  
\date{}

\begin{document}
\maketitle
\setlength{\abovedisplayskip}{3pt}
\setlength{\belowdisplayskip}{3pt}

\begin{abstract}

Interpreting the inner workings of neural models is a key step in ensuring the robustness and trustworthiness of the models, but work on neural network interpretability typically faces a trade-off: either the models are too constrained to be very useful, or the solutions found by the models are too complex to interpret. 
We propose a novel strategy for achieving interpretability that -- in our experiments  -- avoids this trade-off. Our approach builds on the success of using probability as the central quantity, such as for instance within the attention mechanism.
In our architecture, DoLFIn (Distributions over Latent Features for Interpretability),
we do no determine beforehand what each feature represents, and features go altogether into an unordered set. Each feature has an associated probability ranging from 0 to 1, weighing its importance for further processing.
We show that, unlike attention and saliency map approaches, this set-up makes it straight-forward to compute the probability with which an input component supports the decision the neural model makes.
To demonstrate the usefulness of the approach, we apply DoLFIn to text classification, and show that DoLFIn not only provides interpretable solutions, but even slightly outperforms the classical CNN and BiLSTM text classifiers on the SST2 and AG-news datasets.

\end{abstract}

%%%%%%%%%%%%%%%%%%%%%%%%%%%%%%%%%%%%%%%%%%%%%%%%%%%%%%%
\section{Introduction}
\vspace{-1mm}

Having insights into how a trained neural network solves a given task is important, in order to build 
more robust, trustworthy, and accurate models \cite{zhou2016learning,gilpin2018explaining,poerner-etal-2018-evaluating,belinkov-etal-2017-evaluating}.
However gaining insights is often challenging 
because of the  large number of parameters and 
the nonlinear dependencies between components of the model. 
One approach to `opening the blackbox' is to use saliency maps  \cite{jacovi-etal-2018-understanding,gupta-schutze-2018-lisa},
to examine which input components (e.g., words) 
are taken into account. Nevertheless salience alone does not tell us how exactly 
salient components contribute to the decision of a model.
For instance, a sentiment-analysis model might mark both ``boring'' and ``wonderful'' in ``This film would be wonderful, if the beginning wasn't so boring" as salient, but saliency scores do not reveal the crucial interaction between all the words in the sentence that ultimately let it express a negative sentiment. 
%For instance, a sentiment-analysis model might mark both ``boring'' and ``wonderful'' in ``The beginning is boring, but the rest is wonderful'' as salient, but it is unclear which sentiment and to what extent they support.
Alternatively, one can analyse the flow of information through the network, by tracking (relevance) gradients backwards \cite{arras-etal-2017-explaining} or  decomposing (forward) contributions to all intermediate quantities \cite{murdoch2018beyond,jumelet-etal-2019-analysing}. These methods address some of the problems of 
saliency maps, but produce results that themselves require
further interpretation. %(i.e., a quantity can be meaningless unless it is put beside other quantities and interpreted using specific techniques). We conjecture that these problems are rooted in how to measure interpretation quantities.
For instance, such a method might quantify the relative contribution of the 3rd word when processing the 7th word in the 2nd layer of a multilayer LSTM. That quantity, however, does not easily translate to an explanation for the final prediction that the model generates.

A third approach looks at the attention mechanism \cite{bahdanau2014neural}, which regulates  
%computing to 
which components a model attends to. 
Visualising attended components is straight-forward \cite{xu2015show,abnar-zuidema-2020-quantifying}
as an attention weight is the probability that the corresponding component 
is taken into account for further computation. 
At a higher level, as in 
\cite{lei-etal-2016-rationalizing,jasmijn_bastings-etal-2019-interpretable}, 
one can compute rationales which are attended pieces of text. 
However, these approaches face some of the same problems as saliency maps; e.g., % that
attention alone does not indicate
how much an input component
supports a category.

\begin{figure}[t!]
    \centering
    \includegraphics[width=\textwidth]{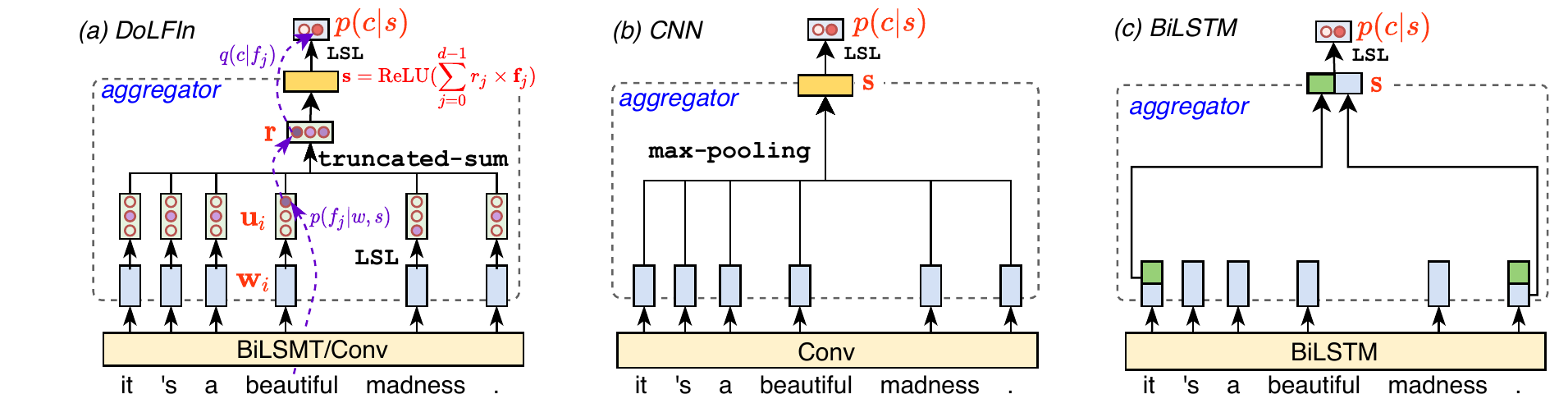}
    \vspace{-5mm}
    \caption{Text classification models. 
    (a) DoLFIn with Bags of latent features (BoLF), using linear-softmax layers (LSL). 
    The whole input text $s$ is mapped to $\mathbf{r}$ which 
    is a bag of latent features. 
    $p(f_i|w,s)$ is the probability of mapping word $w$ (in text $s$) 
    to latent feature $f_i$, 
    and $q(c|f_i)$ is the probability that $f_i$ supports category $c$.
    (b) Traditional CNN proposed by \cite{kim-2014-convolutional}.
    (c) BiLSTM.}
    \label{fig:model}
    \vspace{-5mm}
\end{figure}

In this paper we focus, like the attention-based approach, on probability. We start from the observation that probability is a well studied mathematical tool, that has already been much used in building
explainable and robust neural networks (although we also note that simply turning interpretation quantities from other approaches into probabilities by normalizing is not necessarily helpful, as
these probabilities are not faithful to the 
true behaviour of the model). 
%In this paper, 
We propose DoLFIn (distributions over latent features for interpretability) based on 
probability; our architecture %for building explainable models. 
maps an input (e.g., a text) to a \emph{bag of latent features} (BoLF, see \cref{fig:model}a and \cref{sec:BoLF}) so that interpretation quantities are probabilistic. 
In other words, the representation of an input is a vector whose elements range from 0 to 1, indicating to what extent a latent feature is in the bag. 
To do so, we employ linear-softmax layers 
(i.e., neural layers with softmax activation, LSL for short) 
to map input components to distributions over latent features, and a truncated sum to aggregate the 
resulting distributions.
%\footnote{Using sigmoid might result in a similar approach.}
With this new type of representations, we can easily compute $q(c|w,s)$, 
the probability 
that input component $w$ in context $s$ supports category $c$, by 
decomposing the term to 
$p(f|w,s)$, the distribution over latent features, and 
$q(c|f)$, the probability that feature $f$ supports category $c$. 
The former is given by an LSL and the latter can be estimated by the
input-output statistics. 

To illustrate the feasibility and benefits of this idea, we employ DoLFIn for text classification
that, going beyond saliency maps and attention, 
can tell us the probability a word
supports a category.
%In the example above, BoLF will rely on ``boring'' for category \emph{negative}, 
%and ``wonderful'' for \emph{positive}.
We demonstrate that DoLFIn does not trade off interpretability against classification accuracy. 
We build DoLFIn-conv and DoLFIn-bilstm which are variants of the classical 
CNN proposed by \cite{kim-2014-convolutional}
and BiLSTM text classifier (\cref{fig:model}b,c).
Carrying out experiments on three popular datasets, TREC Question, 
SST2, and AG-news, 
we find that DoLFIn achieves slightly higher accuracy 
than CNN and BiLSTM on SST2 and AG-news. 
It is worth noting that, although used for text classification
in this paper, 
DoLFIn is applicable to a wide range of 
classification tasks 
such as natural language inference and relation prediction.

%%%%%%%%%%%%%%%%%%%%%%%%%%%%%%%%%%%%%%%%%%%%%%%%%%%%%%%
\section{Text Classification Baselines}
\vspace{-1mm}
Given a text of $n$ words $s = (w_0,...,w_{n-1})$ 
and a set of $m$ categories $C = \{ c_0,...,c_{m-1}\}$,
a probabilistic text classifier is $p(c | s)$,
which assigns a probability to the prediction that $c$
is the category of $s$. 
A traditional text classification architecture adopts 
the diagram
\[
s \rightarrow\greybox{encoder} \xrightarrow[]{\mathbf{s}}  \greybox{classifier}\rightarrow
p(c|s)
\] 
where $\mathbf{s} \in \mathbb{R}^{d_s}$ is a vector representing text 
$s$. The classifier module is often a linear-softmax layer. 

Shown in \cref{fig:model}b, 
a classical CNN text classifier~\cite{kim-2014-convolutional} utilises 
a convolutional layer to map  
each word $w_i$ (with its context) to a vector 
$\mathbf{w}_i \in \mathbb{R}^{d_w}$. It then uses max pooling over $\{\mathbf{w}_i\}_{i=0}^{n-1}$ to 
compute $\mathbf{s}$ for text $s$.
A BiLSTM text classifier (\cref{fig:model}c) uses an BiLSTM to compute $\mathbf{w}_i$, 
and an aggregator 
concatenating the backward part of $\mathbf{w}_0$ and 
the forward part of $\mathbf{w}_{n-1}$ to form $\mathbf{s}$.

Training a text classifier is to minimise the cross-entropy loss
\[
L(\theta) = -\frac{1}{|\mathcal{D}_{\text{train}}|} \sum_{(s,c) \in \mathcal{D}_{\text{train}}} \log p(c|s; \theta)
\]
where $\theta$ is the parameter vector and $\mathcal{D}_{\text{train}}$ is a training 
set of $s, c$ pairs.

%%%%%%%%%%%%%%%%%%%%%%%%%%%%%%%%%%%%%%%%%%%%%%%%%%%%%%%%%%%
\section{DoLFIn with Bags of latent features (BoLF)}
\label{sec:BoLF}
\vspace{-1mm}
For simplicity, we introduce DoLFIn as a neural text classifier, 
depicted in \cref{fig:model}a, but DoLFIn should be applicable to several 
classification tasks. 

The key part of DoLFIn, BoLF, is an aggregator consisting of linear-softmax layers 
(LSL) and a truncated sum. 
This aggregator firstly maps each 
$\mathbf{w}_i$ to a distribution $\mathbf{u}_i$ 
over $d$ latent features using LSLs, i.e., $\mathbf{u}_i = \text{LSL}(\mathbf{w}_i)$, so that $\mathbf{u}_{i,j} = p(f_j|w_i,s)$. 
It then uses the truncated sum to compute 
\[
\mathbf{r} = \min(\mathbf{1}, \sum_{i=0}^{n-1} \mathbf{u}_i) \in \mathbb{R}^d
\] 
We apply the element-wise $\min$ operator so that the $j$-th 
entry of $\mathbf{r}$, i.e., $r_j \in [0, 1]$, can be considered 
as a soft indicator of the extent to which the $j$-th latent feature $f_j$ 
contributes to the final output. 
Intuitively, we represent $s$ by a \emph{bag of latent features}
$\mathbf{r}$. 
If $r_j$ is close to 1, feature $f_j$ likely appears in the bag.

Finally, we compute  
\[
\mathbf{s} = \text{ReLU}(\sum_{j=0}^{d-1} r_j \times \mathbf{f}_j )
\]
to represent $s$,  
where each feature $f_j$ is represented by a vector 
$\mathbf{f}_j \in \mathbb{R}^{d_s}$. The sum inside the brackets
can be seen as a bag of the latent feature representations.

\subsection*{Interpretability}
\label{sec:interpretation}

We now show how to analyse the impact of each input component $w$ in context $s$
to the classification decision of the model.
To do so, 
we examine the probability $q(c|w, s)$, which can be seen as the 
support of $w$ in context $s$ for category $c$. 
We decompose this probability by (see the purple arrows in \cref{fig:model}):
\begin{equation}
q(c | w, s) = \sum_{j=0}^{d-1} q(c | f_j) p(f_j | w, s) 
\label{eq:interpretability}
\end{equation}
where $p(f_j|w,s)$ are given by the used LSLs as mentioned above.
(Note that, because we aggregate $p(f_j | w, s)\;\forall j$ into $\mathbf{r}$, 
we do not need to take $\mathbf{r}$ into this equation.) 

Because directly computing $q(c|f)$ is not trivial, we 
approximate it using the statistics of the model's input-output. 
Recall that the latent features from 
$s$'s words are aggregated into 
$\mathbf{r}$, so that if $r_j$ is close to 1, $f_j$ is on and used 
to make the prediction. For simplicity, we assume that 
$f_j$ is on when $r_j > \delta$ for $\delta \in [0, 1]$. 
Let $S$ be a large set of unlabelled texts. 
Let 
$\text{count}_S(c, f_j)$ be the number of texts $s \in S$, 
whose $r_j > \delta$ and which are assigned to category $c$ by the model. Then
\[
q(c | f_j) \approx \frac{\text{count}_S(c, f_j)}{\sum_{c'} \text{count}_S(c', f_j)}
\]
Intuitively, we take into account how many times feature 
$f_j$ appears to yield the prediction $c$. 
Consequently, the closer $q(c | f_j)$ is to 1, the more likely $f_j$ 
supports category $c$.
Besides, if $q(.|f_j)$ is close to uniform, $f_j$ is not helpful
for classification. 
In our experiments, we set $\delta = 0.5$ and $S$ the texts of 
dev sets.

%%%%%%%%%%%%%%%%%%%%%%%%%%%%%%%%%%%%%%%%%%%%%%%%%%%%%%%
\section{Experiments}

In our experimental evaluation we investigate whether DoLFIn can indeed produce interpretable solutions, without sacrificing accuracy.
Our implementation is in Python with Pytorch~\cite{paszke2019pytorch}.
%and AllenNLP~\cite{Gardner2017AllenNLP}.
The source code and data are available at \url{https://github.com/lephong/dolfin}. 
Extra information is provided in the appendix. 
%
%\textbf{Goal:} It is worth recalling that the main goal 
%of DoLFIn is interpretability without sacrificing the classification 
%accuracy. 
%Therefore, we will show that BoLF's decisions are straight-forward 
%to explain, yet its accuracy is on par with those of the CNN and 
%BiLSTM classifiers. 

\begin{table}[t!]
    \small
    \centering
    \begin{tabular}{c|c|c|c|c|c}
        Datasets    &  \#c  & Train & Dev   & Test  & AvgL\\
        \hline
        TREC        & 6     & 5k  & 452   & 500   & 7.5 \\
        SST2        & 2     & 76.9k & 872   & 1821  & 19.2 \\
        AG-news     & 4     & 110k & 10k & 7.6k & 42.3 \\
    \end{tabular}
    \hspace{2mm}
    \scalebox{0.9}{
    \setlength{\tabcolsep}{3pt}
    \begin{tabular}{lcccc}
    & \multicolumn{2}{c}{Conv} & \multicolumn{2}{c}{BiLSTM} \\
    \cmidrule{2-3} \cmidrule{4-5}
    & CNN & DoLFIn-conv & BiLSTM & DoLFIn-bilstm \\
    \midrule
    TREC & \textbf{92.44} \textpm 0.55 & 92.10 \textpm 0.70 & \textbf{92.44} \textpm 0.59 & 91.72 \textpm 0.76 \\
    SST2 & 85.03 \textpm 0.43 & \textbf{85.90} \textpm 0.57 & 86.44 \textpm 0.73 & \textbf{87.23} \textpm 0.66 \\
    AG-news & 92.17 \textpm 0.19 & \textbf{92.59} \textpm 0.17 & 93.26 \textpm 0.11 & \textbf{93.36} \textpm 0.19
    \end{tabular}
    }
    \vspace{-1mm}
    \caption{\emph{Left} - The statistics of TREC,
    SST2, and AG-news datasets. \#c is the number of categories, AvgL 
    is the average length (in words) of test texts.
    \emph{Right} - Accuracy (\%) of the four models 
    on TREC, SST2, and AG-news. 
    We show the mean and standard deviation across 
    five runs.}
    \label{tab:stats-results}
    \vspace{-2mm}
\end{table}

%\subsection{Performance}
\paragraph{Dataset}

We used three following text classification datasets, 
whose statistics are given in \cref{tab:stats-results}-left. 
\begin{itemize}
\item{TREC Question} \cite{li-roth-2002-learning} (TREC for short) is for 
classifying questions into six categories: ABBR (Abbreviation), 
DECS (Description), ENTY (Entity), HUM (Human), LOC (Location), and NUM (Number).
The questions are generally short (7.5 words, on average), such as
``Who are cartoondom 's Super Six ?''

\item{SST2} \cite{socher-etal-2013-recursive} is for predicting the binary sentiment
(positive/negative) of movie reviews. Different from the dev and test sets, 
the train set contains labelled phrases and sentences, rather than sentences alone. 

\item{AG-news} \cite{zhang2015character} is a news topic classification dataset with four topics: WORLD, SPORTS, BUSINESS, and SCI-TECH. Among the three datasets, 
AG-news is the largest in terms of the number of texts 
and the average length.
\end{itemize}

\paragraph{Models}
We evaluated four models CNN, BiLSTM, and DoLFIn-conv/bilstm.
Most of their hyper-parameters are identical to those used in \cite{kim-2014-convolutional} (see Appendix A).
The number of latent features $d$ is 20, 10, and 100 
for DoLFIn when tested on TREC, SST2, and AG-news respectively. 
We used Glove word-embeddings \cite{pennington-etal-2014-glove} 
and Adam optimizer~\cite{kingma2014adam} 
with the default learning rate 0.001. 

\paragraph{Results}
\cref{tab:stats-results}-right shows the results. 
Although DoLFIn performs worse than CNN and BiLSTM on TREC,
it slightly outperforms them on 
SST2 and AG-news. These results suggest that 
using DoLFIn does not sacrifice the classification accuracy.

%\subsection{Interpretation}

\paragraph{Interpretation}

\begin{figure}[t!]
    \centering
    \includegraphics[width=0.5\textwidth]{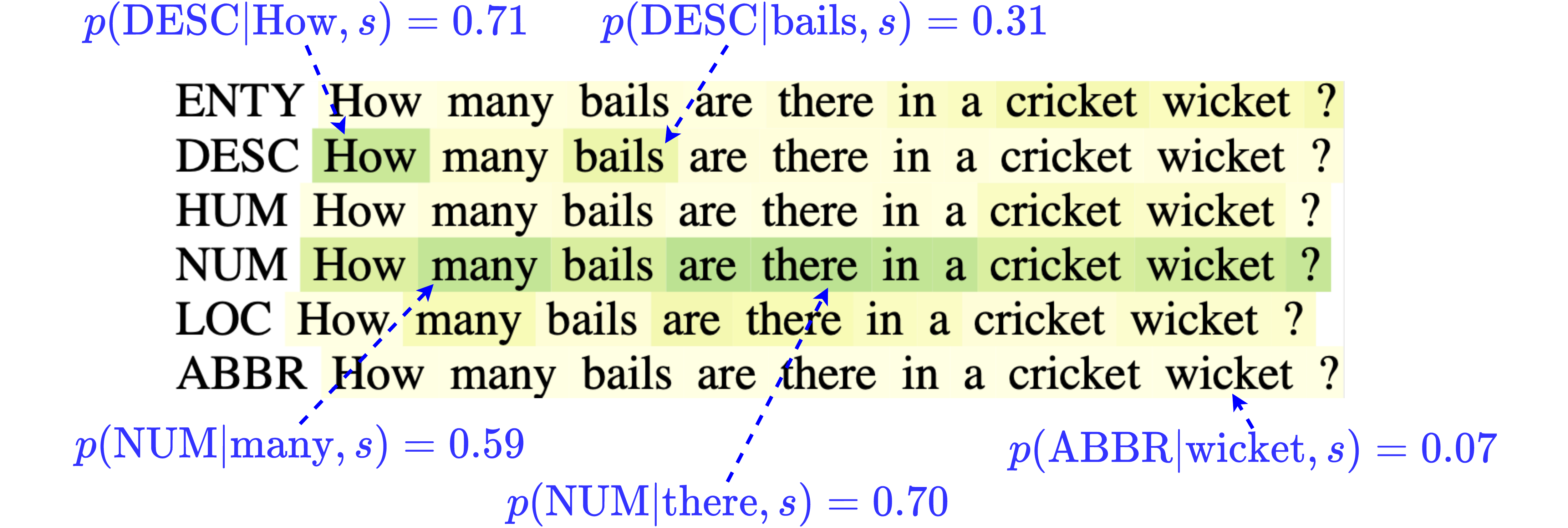}

    \vspace{4mm}
    
    \includegraphics[width=\textwidth]{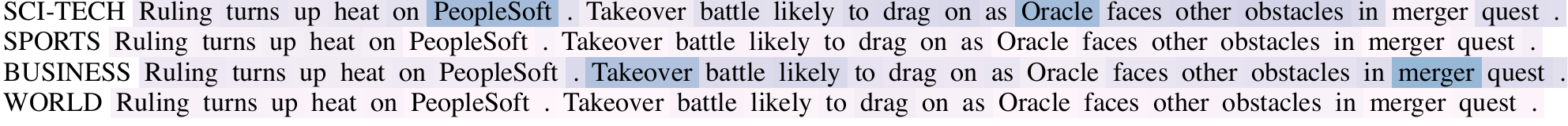}

    \caption{\emph{Top} - A TREC question. Each word is highlighted 
    according to $q(c|w, s)$. For example, ``How'' is highlighted more than ``bails'' in the second question 
    because $q(\text{DESC}|\text{How}, s) > q(\text{DESC}|\text{bails}, s)$.
    The left-most token on each line is the label of category $c$ (e.g. the second question is assigned to category DESC).
    \emph{Bottom} - A text in AG-news.}
    \label{fig:interpret}
\end{figure}

To illustrate how to interpret DoLFIn, 
in \cref{fig:interpret}-top
we visualise a TREC question in the dev set, where
the weight for a word is $q(c|w,s)$, given by \cref{eq:interpretability}. 
If a word (in a context) 
supports the category in question, it will be highlighted. For 
instance, when considering category DESC, we can see that word 
``How'' supports it strongly, ``bails'' slightly, and the other words 
do not. For NUM, ``many'' and ``are there in'' have 
high $q(\text{NUM}|w,s)$. DoLFIn correctly chose NUM.
\cref{fig:interpret}-bottom shows a text in AG-news. DoLFIn reasonably 
focused on ``PeopleSoft'' and ``Oracle'' for SCI-TECH, and 
``Takeover'' and ``merger'' for BUSINESS. It then chose BUSINESS, whereas the
correct topic is SCI-TECH. (This is a difficult case even 
to humans.)

\begin{figure}
    \centering
    \begin{minipage}[b]{.35\textwidth}
    \includegraphics[width=\textwidth]{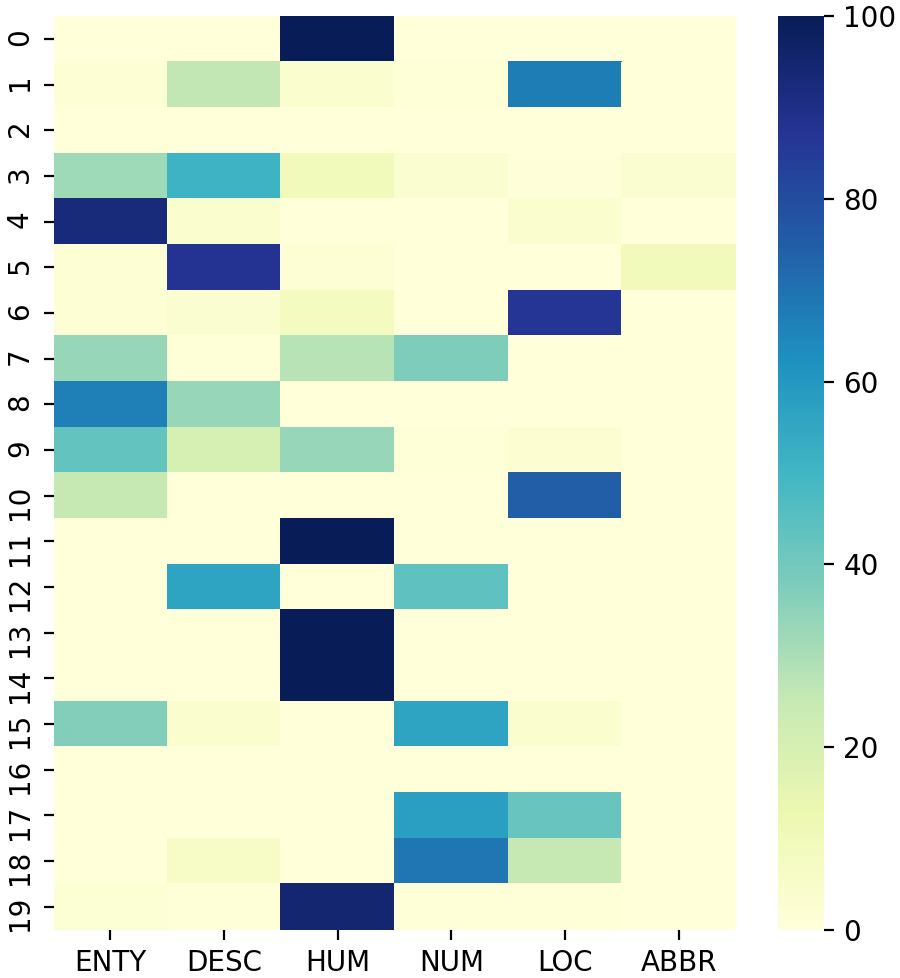}
        \end{minipage}
    \begin{minipage}[b]{.55\textwidth}
        \includegraphics[width=\textwidth]{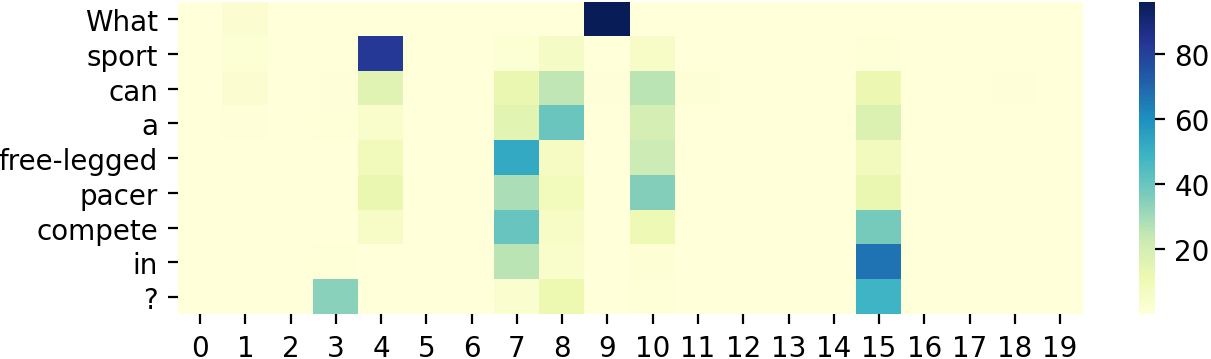}
        \includegraphics[width=\textwidth]{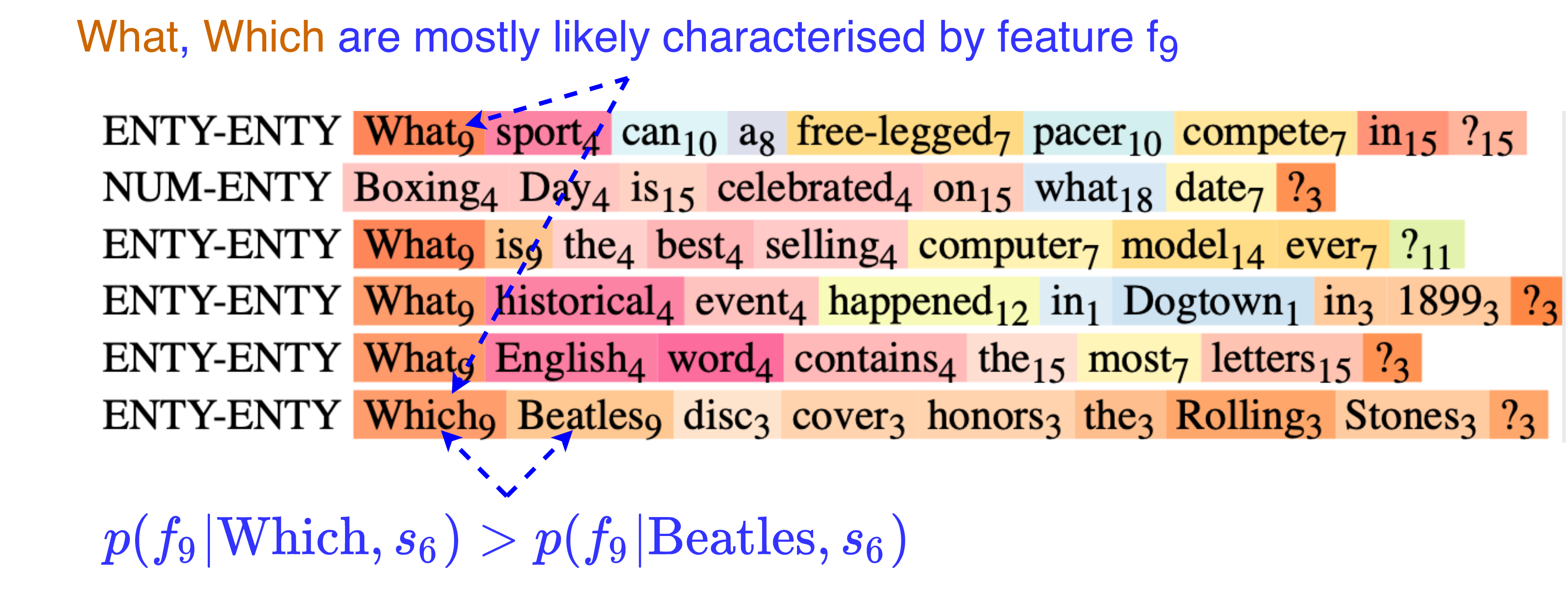}
    \end{minipage}
    \caption{
    \emph{Left} - Heatmap of $q(c|f)$ with 20 latent features 
    for TREC.
    The (j+1)-th row indicates $q(c|f_j)$ (\%). For example, 
    feature $f_9$ supports categories ENTY, DESC, and HUM.
    \emph{Right-top} - Heatmap of $q(f|w,s)$ for A TREC question.
    ``What'' in this sentence is mostly likely 
    characterised by feature $f_9$,
    whereas ``a'' can be $f_7, f_8, f_{10}$, or $f_{15}$.
    \emph{Right-bottom} - TREC questions predicted as ENTY.
    Each word $w$ has a subscript  
    $j = \arg\max_k p(f_k|w, s)$, and is highlighted 
    according to $p(f_j|w, s)$. 
    The left-most token on each line is the gold label-the predicted label.}
    \label{fig:heatmap}
\end{figure}

Another way to interpret the behaviour of DoLFIn is to examine $q(c|f)$ 
and $p(f|w,s)$, especially the meanings of latent features.
\cref{fig:heatmap}-left shows a heatmap visualising $q(c|f)$
for TREC. We can see  
that features $f_2$ and $f_{16}$ are not helpful 
because they support all categories almost uniformly. 
Feature $f_4$ strongly supports ENTY whereas 
$f_0, f_{11}, f_{19}$ support HUM.
Feature $f_9$ prefers ENTY but it can also be used for DESC and HUM.
Interestingly, there are no features strongly
supporting ABBR. DoLFIn seems to rely on the absence of all 
features when predicting ABBR.
\cref{fig:heatmap}-right-top shows a heatmap of $p(f|w,s)$.

\cref{fig:heatmap}-right-bottom
shows TREC questions whose words with their most 
probable latent features are highlighted.\footnote{This example
is cherry-picked for a simple analysis. Deeply analysing the meaning
of latent features is left for future work.}
For instance, the first ``What'' is mapped to feature $f_9$ and 
$q(f_9 | \text{What},s)$ is high (see the first row in 
\cref{fig:heatmap}-right-top). In general, DoLFIn uses $f_9$ 
for words ``What'', ``Which'' that are often for ENTY, 
but sometimes for HUM (e.g., 
``What is the most popular last name~?''), and 
DESC (``What is Java~?'') (see \cref{fig:heatmap}-left). 
If DoLFIn can utilise the next words to choose 
ENTY, it will assign $f_4$ to them (e.g.,
DoLFIn knows that the term ``What \emph{sports}$_4$'' of the first question 
in \cref{fig:heatmap}-right-bottom is for asking about an entity).
Otherwise, it will again use $f_9$
(e.g., DoLFIn still can not decide if ``What \emph{is}$_9$'' of the third question is 
for asking about an entity or a description).

%%%%%%%%%%%%%%%%%%%%%%%%%%%%%%%%%%%%%%%%%%%%%%%%%%%%%%
\begin{comment}
\section{Related Work}
The closest work to ours is 
CNN class activation mapping (CAM) \cite{zhou2016learning}. 
It computes per-category importance scores indicating how much components 
contribute to classification scores. 
Therefore, like DoLFIn, it can shows (i) which input components support a specific category and 
(ii) how they impact 
the classification decision of the model. 
Unlike DoLFIn, 
CAM has a strict structure:
a pooling operation (after a sequence of convolutional layers) 
is connected directly to a linear-softmax output layer.
Consequently, CAM limits the power of 
deep neural networks, whereas DoLFIn does not.  
Besides, DoLFIn can (softly) group components into task-specific clusters 
each of which is represented by a latent feature. In this way, 
we can analyse the similarity between components for a specific task, according to DoLFIn.
\end{comment}

%%%%%%%%%%%%%%%%%%%%%%%%%%%%%%%%%%%%%%%%%%%%%%%%%%%%%%%
\section{Conclusion}

We have proposed a new architecture DoLFIN
based on probability for building explainable models. 
DoLFIn represents input by a \emph{bag of 
latent features} using linear-softmax layers to map input 
components to distributions over latent features, and a truncated 
sum to aggregate these resulting distributions. 
We showed that, different from attention and saliency maps, 
it is straight-forward to compute 
how much an input component supports a category. 
Demonstrating our idea, we applied DoLFIn to text classification. 
Compared with the classical CNN and BiLSTM text classifiers, 
DoLFIn achieved comparable accuracies, but much better interpretability.

%slightly higher accuracy on SST2 and AG-news datasets.

\section*{Acknowledgement}
We would like to thank anonymous reviewers and Thy Tran for
their suggestions and comments.

\newpage
\bibliographystyle{coling}
\bibliography{anthology,ref}

%%%%%%%%%%%%%%%%%%%%%%%%%%%%%%%%%%%%%%%%%%%%%%%%%%%%%%

\appendix
\section{Experiment Setting}

\subsection{Dataset}
The links to the used three dataset are given below.
\begin{itemize}[nosep,noitemsep,leftmargin=*]
    \item TREC Question, \url{https://cogcomp.seas.upenn.edu/Data/QA/QC/}
    \item SST2, \url{https://nlp.stanford.edu/sentiment/}
    \item AG-news, \url{https://github.com/mhjabreel/CharCnn_Keras/tree/master/data/} 
\end{itemize}
The only pre-processing step we applied is tokenization. 

\subsection{Hyper-parameters}
The hyper-parameters of the four models are shown in \cref{tab:hyperparams}.
For DoLFIn, we set the number of latent features $d$ to 20, 10, and 100 
when testing it on TREC, SST2, and 
AG-news respectively. 
We used Glove word-embeddings downloaded from \url{http://nlp.stanford.edu/data/glove.840B.300d.zip} \cite{pennington-etal-2014-glove}.

Following \cite{kim-2014-convolutional}, we applied a dropout layer 
to text representation $\mathbf{s}$, with dropout rate 0.5.

\begin{table}[ht!]
    \centering
    \begin{tabular}{l|l}
        word embedding dimensions $d_w$ & 300 (GloVE)\\
        \hline
        convolutional filter sizes & (3,4,5) \\
        convolutional filter number & 100 \\
        bilstm hidden dimensions & 100 \\
        \hline
        text vector dimenions $d_s$ & 100 (CNN) \\
                                        & 100 (DoLFIn) \\
                                        & 200 (BiLSTM) \\
        \hline 
        minibatch size & 50 \\
        patience (for early stopping) & 10 \\
        optimiser & Adam \\
        learning rate & 0.001 \\
    \end{tabular}
    \caption{Hyper-parameters}
    \label{tab:hyperparams}
\end{table}

\begin{comment}
\section{Visualising $q(c|f)$}
\cref{fig:p_c_f} visualizes the $q(c|f)$ computed by DoLFIn-bilstm 
on TREC Question dataset. 
The $i$-th row indicates $q(c|f_{i-1})$ (\%) for 
$c \in \{ \text{ENTY}, \text{DESC}, \text{HUM}, \text{NUM}, \text{LOC}, \text{ABBR}\}$.
We can see that features $f_2$ and $f_{16}$ are not helpful 
because they support all categories almost uniformly. 
Feature $f_4$ strongly supports ENTY whereas 
$f_0, f_{11}, f_{13}, f_{14}, f_{19}$ support HUM.
Feature $f_9$ prefers ENTY but it can also used for DESC and HUM.

It is worth noticing that there are no features strongly
supporting ABBR. Thus DoLFIn seems to rely on the absence of all 
features when predicting ABBR.

\begin{figure}
    \centering
    \includegraphics[width=0.5\textwidth]{p_c_f.png}
    \caption{Headmap of $q(c|f)$ computed by DoLFIn-bilstm on TREC. 
    The $i$-th row indicates $q(c|f_{i-1})$ (\%) for $c \in \{ \text{ENTY}, \text{DESC}, \text{HUM}, \text{NUM}, \text{LOC}, \text{ABBR}\}$.}
    \label{fig:p_c_f}
\end{figure}
\end{comment}

\end{document}